\documentclass[10pt,twocolumn,letterpaper]{article}

\usepackage[pagenumbers]{cvpr} % To force page numbers, e.g. for an arXiv version
\newcommand{\ours}{\texttt{ICLA}\xspace}
\usepackage[ruled,vlined,linesnumbered]{algorithm2e}
\definecolor{cvprblue}{rgb}{0.21,0.49,0.74}
\usepackage[pagebackref,breaklinks,colorlinks,allcolors=cvprblue]{hyperref}
\usepackage[table]{xcolor}

\def\paperID{####} % *** Enter the Paper ID here
\def\confName{CVPR}

\title{Self-Correction Inside the Model: Leveraging Layer Attention to Mitigate Hallucinations in Large Vision–Language Models}

\author{April Fu\\
Independent Researcher\\
{\tt\small fq.april@gmail.com}
}

\begin{document}
\maketitle

\begin{abstract}
Although Large Vision-Language Models (LVLMs) have made substantial progress, hallucination, where generated text is not grounded in the visual input, remains a challenge. As LVLMs become stronger, previously reported hallucination patterns, such as linguistic bias and ``overthinking'' phenomenon, become far less consistent, making the corresponding mitigation techniques substantially less effective. In this paper, we introduce an \textbf{\underline{I}}nternal self-\textbf{\underline{C}}orrection mechanism utilizing  \textbf{\underline{L}}ayer \textbf{\underline{A}}ttention (\textbf{\ours}) that operates directly on hidden states during generation. Each layer selectively retrieves information from all preceding layers through a diagonal cross-layer attention mechanism, enabling self-refinement without any external correction signals. With introducing and training only 0.2M and 0.1M additional parameters on LLaVA1.5-7B and Qwen2.5-VL-7B, \ours consistently improves visual grounding across multiple hallucination benchmarks, demonstrating its effectiveness for more advanced LVLMs.
\end{abstract}

\section{Introduction}
\label{sec:intro}

Large Vision-Language Models (LVLMs)~\cite{alayrac2022flamingo,liu2023visual,openai2023gpt4,zhu2023minigpt4,li2023blip2} have significantly advanced the capabilities of multimodal learning in broad tasks that require joint reasoning over textual and visual information, including image captioning and visual question answering (VQA)~\cite{lin2014microsoft,goyal2017making,fu2023mme,yu2023mm}. Despite these advancements, hallucination remains a persistent challenge in most LVLMs. This phenomenon refers to cases where the generated text is not grounded in the visual input, often describing objects, or relationships that do not exist in the image~\cite{zhang2023vcd,li2023opera}.

Several studies have investigated the causes of hallucination in LVLMs. One major factor is modality imbalance, where the model tends to over-rely on linguistic priors while underutilizing visual evidence—particularly when images are ambiguous, cluttered, or lack informative content~\cite{zhang2023vcd,wang2023rlhf,zhou2024ibd,li2024povid}. In addition, LVLMs often exhibit a phenomenon termed as ``overthinking'', in which the model captures the correct information in early layers but progressively suppresses visual cues in deeper layers. This over-processing can inadvertently give rise to hallucinated content during text generation~\cite{li2023opera,chuang2023dola,chen2024deco,wang2025damo}.

\begin{figure}
    \centering
    \includegraphics[width=0.99\linewidth]{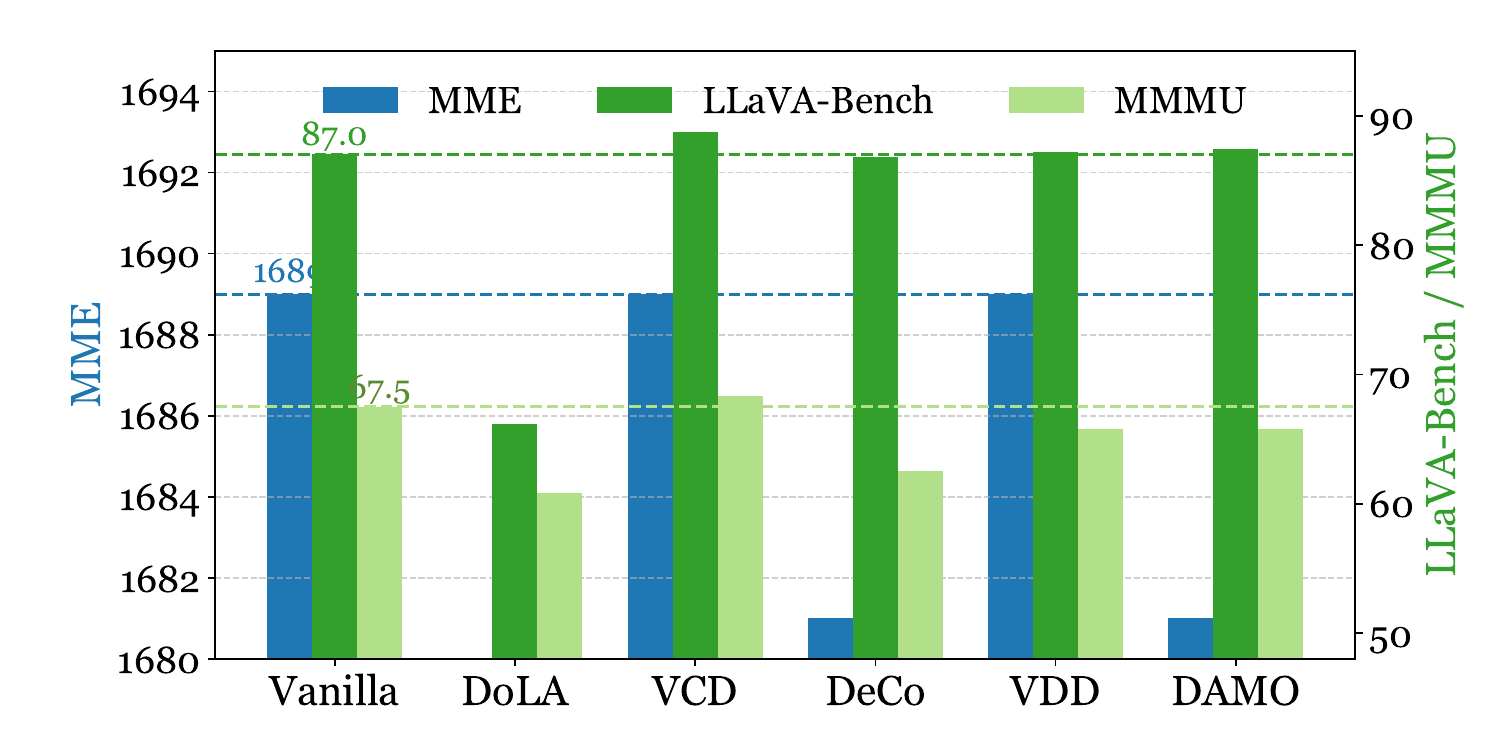}
    \caption{Performance degradation of existing methods on Qwen2.5-VL-7B, where DoLA exhibits a significant drop, while the performance drop of MME exceeds the plot range and is not shown in the figure.}
    \label{fig:performance-drop}
\end{figure}

Many approaches have been proposed to address this limitation. Training-based methods aim to enhance grounding by leveraging supervised fine-tuning (SFT) or reinforcement learning from human feedback (RLHF) on high-quality data~\cite{wang2023rlhf,li2024povid}. Prompt engineering strategies, such as self-correction~\cite{kumar2024training,wu2024large} and self-consistency~\cite{wang2022self,chen2023universal}, further improve model reliability by iteratively refining outputs or aggregating multiple generations to reduce hallucinations. More recently, contrastive decoding (CD) has emerged as an effective strategy, improving grounding by first generating referring logits conditioned on hallucinated visual or textual inputs, and then contrasting the hallucinated logits from the logits with normal input to obtain the final logits~\cite{zhang2023vcd,zhang2024debiasing,zhou2024ibd,lee2024delve,manevich2024mitigating,wang2024mitigating,park2025convis,suo2025octopus}. In addition, several studies have sought to mitigate the ``overthinking'' phenomenon through accumulative decoding (AD), which propagates information from earlier layers that retain visual semantics to later layers, thereby reducing the emergence of hallucinated content~\cite{chen2024deco,wang2025damo,tang2025mitigating,yu2025hallurnn}.

However, with the advancement of training strategies and the availability of high-quality multimodal data, we find that these previously observed hallucination patterns, such as over-reliance on linguistic priors and the ``overthinking'' phenomenon, are no longer clearly present in recent LVLMs. Through an in-depth analysis of a large number of erroneous cases from Qwen2.5-VL-7B~\cite{bai2025qwen2}, including the inspection of its internal probability dynamics during inference, we observe \textit{no consistent or significant hallucination trends} as reported in earlier works.
Moreover, as shown in Figure~\ref{fig:performance-drop}, when evaluating representative methods designed to mitigate these issues on Qwen2.5-VL-7B, we observe that most of them lead to a noticeable performance drop, while only a few perform comparably to the vanilla model, bringing no observable improvements overall.

These observations motivate us to design a more adaptive and scalable paradigm for hallucination mitigation in modern advanced LVLMs, independent of any specific hallucination pattern. In this paper, we propose an \underline{{\textbf{I}}}nternal self-\underline{\textbf{C}}orrection mechanism utilizing \underline{\textbf{L}}ayer \underline{\textbf{A}}ttention (\ours) that operates within the model’s hidden states during generation. Specifically, each layer can actively retrieve information from all preceding layers, self-correcting its hidden state representation according to its current context, thereby mitigating potential hallucinations iteratively. We formulate this process as a cross-layer attention operation, where the hidden state of the current layer serves as the query, and those from all preceding layers serve as keys and values. This design makes \ours particularly suitable for addressing hallucinations in more advanced models where no clear patterns are observable.

Furthermore, to prevent information leakage and cross-position contamination, we apply a diagonal attention mask along the token dimension, ensuring that each hidden state at the current layer can only attend to the hidden states of the same position from all preceding layers, and not to any hidden state of other positions.
The aggregated cross-layer attention output information is then integrated back into the hidden state of current layer, refining its representation and reinforcing visual grounding.

We apply \ours on LLaVA1.5-7B and Qwen2.5-VL-7B, where most prior hallucination patterns and method evaluations were conducted on LLaVA1.5-7B, and Qwen2.5-VL-7B represents a more advanced LVLM. By introducing and training only 0.2M and 0.1M additional parameters on LLaVA1.5-7B and Qwen2.5-VL-7B, respectively, \ours achieves strong performance across multiple hallucination benchmarks on both models, demonstrating its effectiveness in mitigating hallucinations. Notably, \ours attains excellent results on Qwen2.5-VL-7B, further highlighting its suitability for complex and advanced LVLMs. The main contributions of this paper are summarized as follows:
\begin{itemize}
\item We reveal that previously observed hallucination patterns and corresponding mitigation methods are no longer effective for more advanced LVLMs.
\item We propose \ours, an internal self-correction mechanism utilizing layer attention, in which each hidden state can adaptively retrieve information from preceding layers and refine itself accordingly.
\item Extensive experiments on LLaVA1.5-7B and Qwen2.5-VL-7B demonstrate the effectiveness of \ours. Notably, \ours achieves state-of-the-art performance on Qwen2.5-VL-7B, highlighting its suitability for complex and advanced LVLMs.
\end{itemize}
\section{Related Work}

\subsection{Causes of Hallucination in LVLMs}
\paragraph{Modality Imbalance.} Modality imbalance refers to the tendency of LVLMs to over-rely on language priors while underutilizing visual information~\cite{li2022contrastive,han2022visual,guan2024hallusionbench,sennrich2023mitigating,wang2024mitigating,kaul2024throne}. This issue primarily stems from the model architecture, where a visual encoder is typically connected to a pre-trained large language model (LLM). As a result, the linguistic component often dominates multimodal reasoning. For instance, VCD~\cite{zhang2023vcd} attributes hallucinations to the strong statistical biases inherent in language models, particularly when visual cues are weak or ambiguous. Similarly, IBD~\cite{zhou2024ibd} observes that LVLMs tend to overlook fine-grained visual details, producing linguistically plausible but visually ungrounded responses.

\paragraph{``Overthinking''.} Beyond modality imbalance, LVLMs also tend to ``overthink''~\cite{chen2024deco,wang2025damo,tang2025mitigating,yu2025hallurnn}, a phenomenon where the model initially inferences the correct information but subsequently modifies or overrides it in later layers. For example, DeCo~\cite{chen2024deco} finds that visual features captured in early layers are progressively suppressed in deeper layers, weakening visual grounding. DAMO~\cite{wang2025damo} further demonstrates that, although LVLMs encode accurate visual cues, unstable activations in later layers reduce the probabilities of correct tokens, leading to hallucinations. Similarly, DCLA~\cite{tang2025mitigating} confirms this trend, showing that hallucinations primarily emerge during the later decoding stages, consistent with DAMO’s findings.

\subsection{Methods of Hallucination Mitigation in LVLMs}

\paragraph{Training-based Methods.}
Training-based approaches aim to improve visual grounding by fine-tuning LVLMs on high-quality data or leveraging reinforcement learning from human feedback (RLHF)~\cite{chuang2023dola,wang2023rlhf,li2024povid,xu2024hio,wang2024mitigating,gekhman2024does,hu2024mitigating}. For instance, RLHF-V~\cite{wang2023rlhf} and POVID~\cite{li2024povid} construct pairs of hallucinated and non-hallucinated outputs to fine-tune models via Direct Preference Optimization (DPO) algorithm~\cite{rafailov2023direct}, penalizing hallucinated generations. HIO~\cite{xu2024hio}, on the other hand, employs a contrastive loss to help the model distinguish grounded visual context from hallucinated text, thereby reinforcing alignment between visual inputs and generated outputs.

\paragraph{Non-Training-based Methods.}
Since training-based methods require substantial computational resources and high-quality curated data for fine-tuning, non-training-based methods have recently gained increasing attention for hallucination mitigation. These approaches primarily focus on analyzing and leveraging the internal probability dynamics of LVLMs to reduce hallucinations. For example, contrastive decoding (CD) contrasts logits between original and perturbed visual or textual inputs to alleviate hallucinations by reducing the model’s over-reliance on linguistic priors~\cite{zhang2023vcd,huo2024self,zhang2024debiasing,lyu2024alleviating,zhou2024ibd,lee2024delve,manevich2024mitigating,wang2024mitigating,park2025convis,suo2025octopus,zhang2025self}. In addition, to address the “overthinking” limitation, accumulative decoding (AD) has been proposed. AD operates on the hidden states during generation in an accumulative manner, complementing visual information from earlier layers to later ones, thereby preserving visual grounding and mitigating the suppression of visual cues in deeper layers~\cite{yu2025hallurnn,tang2025mitigating,wang2025damo,chen2024deco}.

\subsection{Layer Attention}
Layer attention has been explored in earlier studies on deep convolutional neural networks (DCNNs) as a way to improve information flow across depth. For instance, DIANet~\cite{huang2020dianet} employs a shared LSTM to capture inter-layer dependencies, and subsequent works~\cite{claster2025adaptive,wang2024strengthening,fang2023cross} further develop more sophisticated mechanisms for propagating and refining information across layers. These approaches, although primarily designed for smaller models and general feature enhancement, provide valuable insight into the benefits of cross-layer communication. Inspired by this line of research, our work revisits cross-layer interaction in the context of LVLMs, leveraging it for adaptive refinement, thereby iteratively mitigating hallucination.

\section{Method}

We begin by reviewing the information flow in LVLMs to highlight the transformation of hidden states. Subsequently, we introduce the overall architecture of our proposed Internal Self-Correction via Layer Attention (\ours) in detail. Finally, we elaborate the Cross-Layer Attention (CLA) module, which is the key component in \ours.

\begin{figure}
    \centering
    \includegraphics[width=0.99\linewidth]{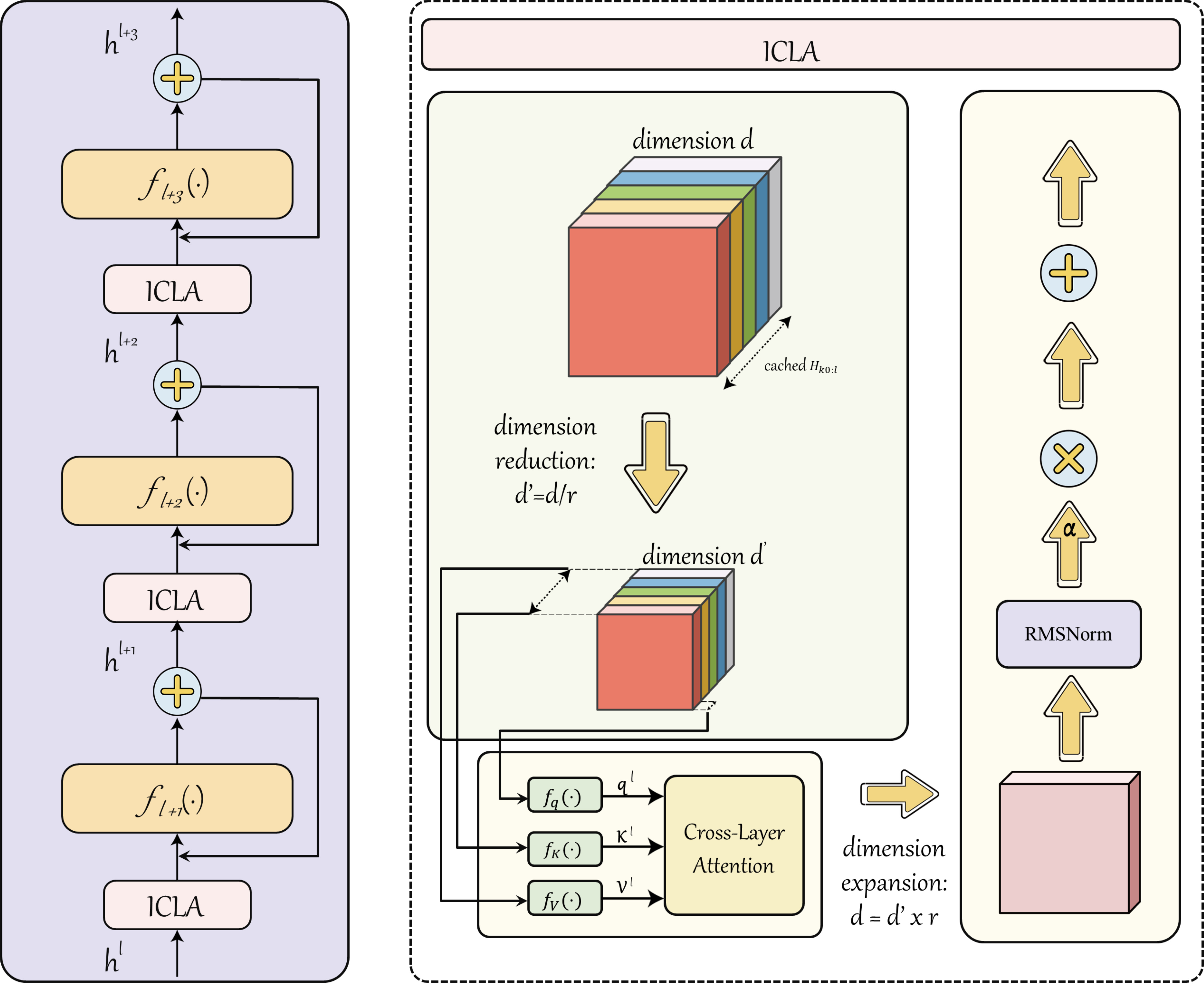}
    \caption{Overall architecture of \ours.}
    \label{fig:arch}
\end{figure}

% \subsection{Preliminaries}
% Large Vision-Language Models (LVLMs) process multimodal inputs through a stack of transformer layers that iteratively refine token representations.  
% Let the multimodal input be $x = (x_1, \dots, x_T)$, where each $x_i$ denotes either a visual or textual token embedding. The model maintains a sequence of hidden states across $L$ layers, defined as:
% \begin{equation}
%     h^l = (h^l_1, \dots, h^l_T), \quad l = 0, \dots, L,
% \end{equation}
% where $h^l_i \in \mathbb{R}^d$ represents the hidden state of token $x_i$ at $l$-th layer.  
% By definition, the initial hidden state $h^0$ corresponds to the input embeddings $x$.

% Each layer updates the hidden states based on:
% \begin{equation}
%     h^l = f_l(h^{l-1})+h^{l-1},
%     \label{equ:layer_update}
% \end{equation}
% where $f_l$ denotes the transformation of the $l$-th layer, typically including self-attention and feed-forward submodules. After the final layer, the hidden states $h^{L}$ are used to predict the next token distribution:
% \begin{equation}
%     P(y_t \mid x, y_{<t}) = \mathrm{Softmax}(W_o h^L_{t-1}),
% \end{equation}
% where $h^L_{t-1} \in \mathbb{R}^d$ represents the final hidden representation of the $t-1$-th token and $W_o$ is the output projection matrix, the $d$ is the dimension of the hidden state.

\subsection{Preliminaries}
LVLMs process multimodal inputs through a stack of transformer layers that iteratively refine token representations.  
Let the multimodal input be $x = (x_1, \dots, x_T)$, where each $x_i$ denotes either a visual or textual token embedding. The model maintains a sequence of hidden states across $L$ layers, defined as:
\begin{equation}
    h^l = (h^l_1, \dots, h^l_T), \quad l = 0, \dots, L,
\end{equation}
where $h^l_i \in \mathbb{R}^d$ represents the hidden state of token $x_i$ at the $l$-th layer, and $d$ denotes the dimensionality of the hidden states.
By definition, the initial hidden state $h^0$ corresponds to the input embeddings $x$.

Within the model, the hidden states are iteratively updated through transformer layers along with the residual connetion:
\begin{equation}
    h^{l+1} = f_{l+1}(h^{l}) + h^{l},
    \label{equ:layer_update}
\end{equation}
where $f_{l+1}(\cdot)$ denotes the transformation function of the $(l+1)$-th layer, typically consisting of multi-head self-attention and feed-forward submodules. For brevity, we omit the normalization here.  
After the final layer, the hidden states $h^{L}$ are used to predict the next-token distribution:
\begin{equation}
    P(y_t \mid x, y_{<t}) = \mathrm{Softmax}(W_o\cdot h^L_t),
\end{equation}
where $h^L_t$ is the final-layer hidden state of the $t$-th token, $W_o$ is the output projection matrix.

\subsection{Internal Self-Correction via Layer Attention}
To address hallucinations in advanced LVLMs, where previously observed patterns and mitigation strategies fail to generalize, we propose \ours, which enables each hidden state to adaptively retrieve informative representations from preceding layers, thereby dynamically and iteratively refining its own representation and self-correcting the potential hallucinations. The overall architecture of \ours is shown in Figure~\ref{fig:arch}.

Specifically, at the $l$-th layer, we first store the hidden states from the $k_0$-th to the $l$-th layer, forming a hidden state cache for the subsequent cross-layer attention mechanism:
\begin{equation}
H_{k_{0:l}} = \{h^{k}\}_{k = k_0}^{l} \in \mathbb{R}^{(l - k_0 + 1) \times T \times d},
\end{equation}
where $k_0$ serves as the starting layer for hidden state refinement as modifying early layers may destroy the normal inference context~\cite{wang2025damo,yu2025hallurnn,tang2025mitigating}. After obtaining the hidden state cache, cross-layer attention is applied to retrieve relevant information from previous layers:
\begin{equation}
    O^l = \mathrm{CLA}(H_{k_{0:l}}),
\end{equation}
where $\mathrm{CLA}$ denotes the cross-layer attention module (detailed in Section~\ref{sec:CLS}) and $O^l \in \mathbb{R}^{T \times d}$ represents the attention output.  
Finally, the attention output is scaled and used to refine the current hidden states:
\begin{equation}
    h^l := h^l + \alpha \cdot \mathrm{RMSNorm}(O^l),
\end{equation}
where $\alpha$ is the scaling factor controlling the refinement strength. Here, $\mathrm{RMSNorm}$ refers to root mean square normalization, which normalizes the hidden state along the feature dimension to stabilize training and preserve the scale of activations. Then the refined hidden states $h^l$ are used to compute the next layer’s hidden states $h^{l+1}$ for iterative update according to Equation~\ref{equ:layer_update}.

\subsection{Cross-Layer Attention}\label{sec:CLS}
In this section, we describe the Cross-Layer Attention (CLA) module, which differs from standard self-attention and cross-attention by processing information across transformer layers rather than within a single layer or modality.

Given the hidden state cache $H_{k_{0:l}}$, we first compute the query, key, and value projections as follows:
\begin{equation}
\left\{
\begin{aligned}
&q^l = W_q \cdot h^l \\
&K^l = W_K \cdot H_{k_{0:l}} \\
&V^l = W_V \cdot H_{k_{0:l}},
\end{aligned}
\right.
\end{equation}
where $h^l$ is the last element in $H_{k_{0:l}}$, representing the current hidden state. $W_q$, $W_K$, and $W_V$ are learnable linear layers with a bottleneck: we first reduce the hidden dimension from $d$ to latent hidden dimension $d' = d / r$ for improving training and inference efficiency and $r$ serves as the reduction ratio, similar to~\cite{huang2020dianet,wang2024strengthening}.

Then, the obtained query, key, and value are used to compute the attention output. To ensure that each token at the $i$-th position in the current $l$-th layer only attends to representations from previous layers at the same $i$-th position, we extract the diagonal of the attention matrix over the token dimension:

\begin{equation}
\left\{
\begin{aligned}
\boldsymbol{\alpha}^{l} &= \text{diag} \left( \text{softmax} \left( \frac{{q}^l \cdot {(K^{l})}^\top}{\sqrt{d'}} \right) \right) \\
O^l &= W_\text{out} \cdot \sum_{k = k_0}^{l} \boldsymbol{\alpha}_k^{l} \cdot V^l_k,
\end{aligned}
\right.
\end{equation}

where $\boldsymbol{\alpha}^{l}$ denotes the attention weights, and $W_\mathrm{out}$ projects the attention output from the latent dimension $d'$ back to the original hidden dimension $d$. This design enables the attention mechanism to be performed in the latent space, efficiently reducing computational cost. The diagonal-only formulation ensures that each token aggregates information vertically across layers without interacting with other token positions.

Notably, the CLA module is parameter-shared across the whole network to reduce the introduced parameters and enhance the training efficiency. The detailed algorithm is shown in Algorithm~\ref{alg:icla}.

% \paragraph{Comparison with Other Layer Attention}
% While conventional layer attention mechanisms have been widely explored in the computer vision domain~\cite{fang2023cross,wang2024strengthening}, their primary focus lies in enhancing feature reuse by propagating low-level visual details to deeper layers. These methods typically operate on concrete image features to facilitate tasks such as object detection or semantic segmentation. In contrast, our \ours applies the concept of layer attention to the abstract hidden states in LVLMs. Instead of modeling spatially grounded visual cues, \ours attends over semantically rich representations across layers during decoding. This demonstrates that the benefits of layer-wise information integration extend beyond visual feature refinement, and are also effective for modeling abstract semantics in multimodal generation tasks.

\begin{algorithm}[t]
\caption{Internal Self-Correction via Layer Attention (\ours)}
\label{alg:icla}
\KwIn{Multimodal input $x = (x_1, \dots, x_T)$; total layers $L$}
\KwOut{Refined hidden states $h^L$}

\BlankLine
\textbf{Initialization:} $h^0 \leftarrow x$ \tcp*[r]{Input embeddings}

\For{$l = 1$ \KwTo $L$}{
    $h^l \leftarrow f_l(h^{l-1})+h^{l-1}$ \tcp*[r]{Standard transformer update}
    
    \If{$l > k_0$}{
        $H_{k_{0:l}} = \{h^{k}\}_{k=k_0}^{l}$ \tcp*[r]{Cache recent hidden states}
        
        \textbf{Cross-Layer Attention:}\\
        \Indp
        $q^l = W_q\cdot h^l$, \quad
        $K^l = W_K\cdot H_{k_{0:l}}$, \quad
        $V^l = W_V\cdot H_{k_{0:l}}$ \\
        $\boldsymbol{\alpha}^{l} = \mathrm{diag}\!\left(\mathrm{softmax}\!\left(\frac{q^l\cdot ({K^{l}})^{\top}}{\sqrt{d'}}\right)\right)$ \\
        $O^l = W_\mathrm{out}\! \cdot\! \sum_{k=k_0}^{l} \boldsymbol{\alpha}_k^{l} \cdot V^l_k$ \\
        \Indm
        $h^l \leftarrow h^l + \alpha \cdot \mathrm{RMSNorm}(O^l)$ \tcp*[r]{Refine current states}
    }
}
\Return{$h^L$}
\end{algorithm}

\definecolor{lightblue}{RGB}{220,235,255}

\begin{table*}[htbp]
\centering
\begin{tabular}{lcccccc}
\toprule
\textbf{Model}& \multicolumn{3}{c}{LLaVA1.5-7B}&\multicolumn{3}{c}{Qwen2.5-VL-7B}\\
\midrule
\textbf{Method}& MME& LLaVA-Bench& MMMU& MME& LLaVA-Bench& MMMU\\
\midrule
Vanilla & 1484 & 59.6 & 35.3 & 1689 & 87.0& 67.5\\
DoLA & 1485 & 60.5 & 35.7 & 1403 & 66.2& 60.8\\
VCD & 1469 & 60.6 & 35.8 & 1689 & 88.7& 68.3\\
DeCo & 1456 & 57.0 & 33.9 & 1681 & 86.8& 62.5\\
POVID & 1483 & 60.2 & 35.3 & - & - & -\\
VDD & 1484 & 59.4 & 34.9 & 1689 & 87.2& 65.8\\
DAMO & 1495 & 57.7 & 34.4 & 1681 & 87.4& 65.8\\
\textbf{\textit{\ours (Ours)}} & \cellcolor{lightblue}\textbf{1499} & \cellcolor{lightblue}\textbf{61.9} & \cellcolor{lightblue}\textbf{35.9} & \cellcolor{lightblue}\textbf{1711} & \cellcolor{lightblue}\textbf{90.2} & \cellcolor{lightblue}\textbf{69.2}\\
\bottomrule
\end{tabular}
\caption{Experimental results on MME (total perception score), LLaVA-Bench (overall accuracy), and MMMU (accuracy) for LLaVA1.5-7B and Qwen2.5-VL-7B. The best results are highlighted in \textbf{bold} with light blue shading.}
\label{tab:main-results-all}
\end{table*}

% \begin{table*}[htbp]
% \centering
% \begin{tabular}{lcccccccc}
% \toprule
% Method & Vanilla &  DoLa & VCD & DeCo & POVID & VDD &  DAMO &\textbf{Ours(CLAM)} \\
% \midrule
% All (\%) &  59.6&  60.5& 60.6& 57.0& 60.2& 59.4&  57.7&\textbf{61.9}\\
% Complex reasoning (\%) &  \textbf{72.1}&  70.5& 70.7& 66.8& 69.4& 71.9&  65.1&70.4\\
% Conversation (\%) &  50.6&  53.7& 55.0& 52.7& 55.0& 52.3&  55.0&\textbf{59.6}\\
% Detail description (\%) &  48.8&  50.0& 49.3& 45.0& \textbf{50.4}& 45.7&  48.0&50.0\\
% \bottomrule
% \end{tabular}
% \caption{Comparison of LLaVA-Bench performance for various baselines using LLaVA1.5-7B. The highest scores in each column are shown in bold.}
% \label{tab:llava-bench}
% \end{table*}
\section{Experiments}

\subsection{Experimental Setup}

% \definecolor{lightblue}{RGB}{220,235,255}

% \begin{table*}[htbp]
% \centering
% \begin{tabular}{lcccccc}
% \toprule
% \textbf{Model}& \multicolumn{3}{c}{LLaVA1.5-7B}&\multicolumn{3}{c}{Qwen2.5-VL-7B}\\
% \midrule
% \textbf{Method}& MME& LLaVA-Bench& MMMU& MME& LLaVA-Bench& MMMU\\
% \midrule
% Vanilla & 1484 & 59.6 & 35.3 & 1689 & 87.0& 67.5\\
% DoLA & 1485 & 60.5 & 35.7 & 1403 & 66.2& 60.8\\
% VCD & 1469 & 60.6 & 35.8 & 1689 & 88.7& 68.3\\
% DeCo & 1456 & 57.0 & 33.9 & 1681 & 86.8& 62.5\\
% POVID & 1483 & 60.2 & 35.3 & - & - & -\\
% VDD & 1484 & 59.4 & 34.9 & 1689 & 87.2& 65.8\\
% DAMO & 1495 & 57.7 & 34.4 & 1681 & 87.4& 65.8\\
% \textbf{\textit{\ours (Ours)}} & \cellcolor{lightblue}\textbf{1499} & \cellcolor{lightblue}\textbf{61.9} & \cellcolor{lightblue}\textbf{35.9} & \cellcolor{lightblue}\textbf{1711} & \cellcolor{lightblue}\textbf{90.2} & \cellcolor{lightblue}\textbf{69.2}\\
% \bottomrule
% \end{tabular}
% \caption{Experimental results on MME (total perception score), LLaVA-Bench (overall accuracy), and MMMU (accuracy) for LLaVA1.5-7B and Qwen2.5-VL-7B. The best results are highlighted in \textbf{bold} with light blue shading.}
% \label{tab:main-results-all}
% \end{table*}

\paragraph{Models and Baselines.}
We implement our \ours on two popular LVLMs—LLaVA1.5-7B and Qwen2.5-VL-7B—for comprehensive evaluation. We compare our \ours with several strong hallucination mitigation baselines.
Vanilla serves as the base model (LLaVA1.5-7B~\cite{liu2023visual} or Qwen2.5-VL-7B~\cite{bai2025qwen2}). VCD~\cite{zhang2023vcd} mitigates hallucinations by contrasting outputs from original and distorted visual inputs. VDD~\cite{zhang2023vcd} extends VCD with post-hoc debiasing and debiased sampling. DoLA~\cite{chuang2023dola} contrasts logits across transformer layers to mitigate hallucinations. POVID~\cite{li2024povid} employs Direct Preference Optimization (DPO) to align the model and reduce hallucinations. DeCo~\cite{chen2024deco} fuses information from selected preceding layers into the final decoding layer. DAMO~\cite{wang2025damo} introduces momentum-based decoding to maintain inter-layer consistency and enhance factual grounding. For all experiments, the temperature is set to 0 for \textit{greedy decoding} to ensure fair comparison.

\vspace{-3mm}
\paragraph{Benchmarks and Metrics.}
We evaluate \ours on four established hallucination benchmarks: POPE~\cite{li2023evaluating}, MME~\cite{fu2023mme}, MMMU~\cite{yue2024mmmu}, and LLaVA-Bench~\cite{liu2023visual}.
POPE measures hallucination resistance on MSCOCO and A-OKVQA datasets under adversarial, random, and popular settings, reporting both accuracy and F1 scores. MME evaluates perception-related hallucinations, using the total perception score. MMMU tests multimodal reasoning ability with official accuracy.
LLaVA-Bench adopts GPT-4o-based evaluation over perception, reasoning, and dialogue, providing both overall scores.

\vspace{-3mm}
\paragraph{Training Details.}
We train \ours using a lightweight tuning strategy with positive samples from the POVID training set, which consists of 17K examples randomly sampled from LLaVA-Instruct-150K. Notably, all data overlap with those used in the official LLaVA training, so no additional knowledge is introduced. All model parameters are frozen except for those in the \ours module. The key hyperparameters are set as default: learning rate $lr=2e-5$, starting layer $k_0 = 16$, reduction ratio $r = 128$, and scaling factor $\alpha = 0.02$, training epochs $epoch=3$.

% \begin{figure}[t]
% \centering
% \includegraphics[width=0.95\columnwidth]{CVPR2026/Figure/MMMU_MME_evaluation.pdf}
% \caption{(a) Comparison between various baselines on MMMU benchmark; (b) Comparison between various baselines on MME-Perception benchmark;}
% \label{fig:mmmu-radar}
% \vspace{-3mm}
% \end{figure}

\begin{table*}[t]
\centering
\small
\resizebox{\textwidth}{!}{
\begin{tabular}{llcccccccc}
\toprule
\multirow{3}{*}{\textbf{Setting}} & \multirow{3}{*}{\textbf{Method}} 
& \multicolumn{4}{c}{\textbf{LLaVA1.5-7B}} 
& \multicolumn{4}{c}{\textbf{Qwen2.5-VL-7B}} \\
\cmidrule(lr){3-6} \cmidrule(lr){7-10}
& & \multicolumn{2}{c}{\textbf{MSCOCO}} & \multicolumn{2}{c}{\textbf{A-OKVQA}} 
& \multicolumn{2}{c}{\textbf{MSCOCO}} & \multicolumn{2}{c}{\textbf{A-OKVQA}} \\
\cmidrule(lr){3-4} \cmidrule(lr){5-6} \cmidrule(lr){7-8} \cmidrule(lr){9-10}
& & F1 & Acc & F1 & Acc & F1 & Acc & F1 & Acc \\
\midrule

\multirow{7}{*}{Adversarial}
& Vanilla & 81.76 & 79.77 & 76.12 & 69.37 & 80.72 & 83.40 & 80.73 & 79.53 \\
& DoLA & 81.69 & 79.73 & 76.22 & 69.53 & 77.44 & 80.63 & 80.91 & 76.57 \\
& VCD & 80.50 & 78.33 & 75.05 & 67.97 & 81.13 & 77.77 & 80.73 & 72.03 \\
& DeCo & 81.14 & 78.37 & 74.93 & 67.03 & 80.86 & 83.47 & 80.93 & 80.10 \\
& POVID & 81.88 & 80.03 & 76.18 & 69.47 & -- & -- & -- & -- \\
& DAMO & 81.65 & 79.53 & 75.96 & 68.97 & 80.84 & 83.47 & 81.09 & 80.30 \\
& \textbf{\textit{\ours (Ours)}} 
& \cellcolor{lightblue}\textbf{81.90} & \cellcolor{lightblue}\textbf{80.13} 
& \cellcolor{lightblue}\textbf{76.32} & \cellcolor{lightblue}\textbf{69.73} 
& \cellcolor{lightblue}\textbf{81.50} & \cellcolor{lightblue}\textbf{83.97} 
& \cellcolor{lightblue}\textbf{81.60} & \cellcolor{lightblue}\textbf{80.93} \\
\midrule

\multirow{7}{*}{Popular}
& Vanilla & 86.82 & 86.23 & 83.22 & 80.30 & 81.03 & 83.73 & 85.32 & 85.83 \\
& DoLA & 86.84 & 86.30 & 83.33 & 80.47 & 77.70 & 80.97 & 84.66 & 83.00 \\
& VCD & 84.80 & 83.97 & 81.02 & 77.43 & 81.02 & 77.67 & 85.48 & 74.37 \\
& DeCo & 85.90 & 84.73 & 81.39 & 77.47 & 81.24 & 83.90 & 85.45 & 86.20 \\
& POVID & 86.78 & 86.23 & 83.32 & 80.53 &- & - & - & - \\ 
& DAMO & 86.84 & 86.20 & 83.04 & 79.97 & 81.25 & 83.90 & 85.56 & 86.27 \\
& \textbf{\textit{\ours (Ours)}} 
& \cellcolor{lightblue}\textbf{86.91} & \cellcolor{lightblue}\textbf{86.43} 
& \cellcolor{lightblue}\textbf{83.39} & \cellcolor{lightblue}\textbf{80.57} 
& \cellcolor{lightblue}\textbf{81.92} & \cellcolor{lightblue}\textbf{84.37} 
& \cellcolor{lightblue}\textbf{86.37} & \cellcolor{lightblue}\textbf{87.07} \\
\midrule

\multirow{7}{*}{Random}
& Vanilla & 89.71 & 89.60 & 88.52 & 87.33 & 81.58 & 84.37 & 87.06 & 87.63 \\
& DoLA & 89.77 & 89.70 & 88.55 & 87.37 & 78.02 & 81.57 & 85.91 & 84.43 \\
& VCD & 87.60 & 87.37 & 86.52 & 85.00 & 81.64 & 79.17 & 87.86 & 80.77 \\
& DeCo & 89.29 & 88.83 & 86.53 & 84.67 & 81.78 & 84.50 & 87.27 & 88.10 \\
& POVID & 89.56 & 89.50 & 88.29 & 87.07 & - & - & - & - \\
& DAMO & 89.69 & 89.53 & 87.93 & 86.53 & 81.83 & 84.57 & 87.32 & 88.13 \\
& \textbf{\textit{\ours (Ours)}} 
& \cellcolor{lightblue}\textbf{89.95} & \cellcolor{lightblue}\textbf{89.93} 
& \cellcolor{lightblue}\textbf{88.80} & \cellcolor{lightblue}\textbf{87.70} 
& \cellcolor{lightblue}\textbf{82.59} & \cellcolor{lightblue}\textbf{85.10} 
& \cellcolor{lightblue}\textbf{88.28} & \cellcolor{lightblue}\textbf{89.03} \\
\bottomrule
\end{tabular}
}
\caption{
\textbf{Results on the POPE benchmark.} 
Comparison between \textbf{LLaVA1.5-7B} and \textbf{Qwen2.5-VL-7B}. 
We report F1 Score (\%) and Accuracy (\%) on the MSCOCO and A-OKVQA datasets under Adversarial, Popular, and Random settings. 
The best results are highlighted in \textbf{bold blue}.
}
\label{tab:llava-qwen-pope}
\end{table*}

\begin{table*}[htbp]
\centering
\begin{tabular}{lccccccc}
\toprule
Method & Vanilla &  DoLA & VCD & DeCo & VDD &  DAMO &\textbf{\ours (Ours)} \\
\midrule
Complex Reasoning (\%) &   95.8&  66.5& \cellcolor{lightblue}\textbf{98.6}& 95.2&  94.8&  96.3&93.3\\
Conversation (\%) &   82.4&  65.0&  86.1& 83.7& 85.0&  82.0&\cellcolor{lightblue}\textbf{89.4}\\
Detail Description (\%) &   76.0&  66.9& 74.4& 75.8&  76.4&  77.7&\cellcolor{lightblue}\textbf{85.5}\\
All (\%) &   87.0&  66.2&  88.7& 86.8& 87.2&  87.4&\cellcolor{lightblue}\textbf{90.2}\\
\bottomrule
\end{tabular}
\caption{Comparison of LLaVA-Bench performance for various baselines with backbone Qwen2.5-VL-7B. The best results are highlighted in \textbf{bold} blue.}
\label{tab:llava-bench-qwen}
\end{table*}

\subsection{Experimental Results}

\paragraph{Results on LLaVA1.5-7B.}
As shown in Table~\ref{tab:main-results-all} and Table~\ref{tab:llava-qwen-pope}, \ours consistently outperforms all other baselines on LLaVA1.5-7B.
On the MME benchmark, \ours achieves a 15-point improvement over the Vanilla LLaVA baseline and further surpasses contrastive decoding methods such as VCD and VDD by 30 and 15 points, respectively. On both LLaVA-Bench and MMMU, \ours also attains the best performance.
In particular, on LLaVA-Bench, \ours reaches an accuracy of 61.9\%, representing a 2.3\% improvement over the Vanilla baseline. Compared to DAMO and DeCo, which also operate on hidden states during inference, \ours further outperforms them by 4.9\% and 4.2\%, respectively. On the POPE benchmark, \ours achieves the highest F1 and accuracy scores across both two datasets (MSCOCO and A-OKVQA) under all three settings, demonstrating that the flexible self-correction of \ours is stronger than the curated methods for observed hallucination patterns.

\vspace{-2mm}
\paragraph{Results on Qwen2.5-VL-7B.}
As shown in Table~\ref{tab:main-results-all}, \ours also achieves outstanding results on Qwen2.5-VL-7B across the MME, LLaVA-Bench, and MMMU benchmarks.
On the MME benchmark, we observe an interesting phenomenon: most baseline methods perform on par with or even worse than the Vanilla Qwen2.5-VL-7B. For example, only VCD and VDD achieve comparable scores to the baseline, while all other methods yield lower results, suggesting that these approaches may not generalize well to more advanced LVLMs.
In contrast, \ours achieves a remarkable 22-point improvement over the Vanilla model, demonstrating both its effectiveness and strong adaptability.  \ours also attains the best performance on the MMMU benchmark, surpassing all baselines.

As shown in Table~\ref{tab:llava-qwen-pope}, \ours further delivers strong results on the POPE benchmark. Specifically, it achieves the best F1 and accuracy across all three settings on the MSCOCO and A-OKVQA datasets. These results collectively demonstrate that \ours is particularly effective and well-suited for more advanced LVLMs such as Qwen2.5-VL-7B.

As shown in Table~\ref{tab:llava-bench-qwen}, we provide a fine-grained comparison across three tasks on LLaVA-Bench using Qwen2.5-7B-VL as the backbone to demonstrate the effectiveness of \ours in generalized hallucination mitigation. From an overall perspective, most baselines achieve performance comparable to vanilla Qwen2.5-VL-7B (87.0\%), whereas \ours unexpectedly reaches 90.2\%, yielding a substantial 3.2\% improvement. Examining the task-level results, \ours delivers particularly strong gains on Conversation and Detailed Description, with improvements of 7\% and 9.5\%, respectively. These significant increases further demonstrate the robustness and general applicability of our method across diverse hallucination scenarios.

\begin{figure*}[t]
\centering
\includegraphics[width=2.0\columnwidth]{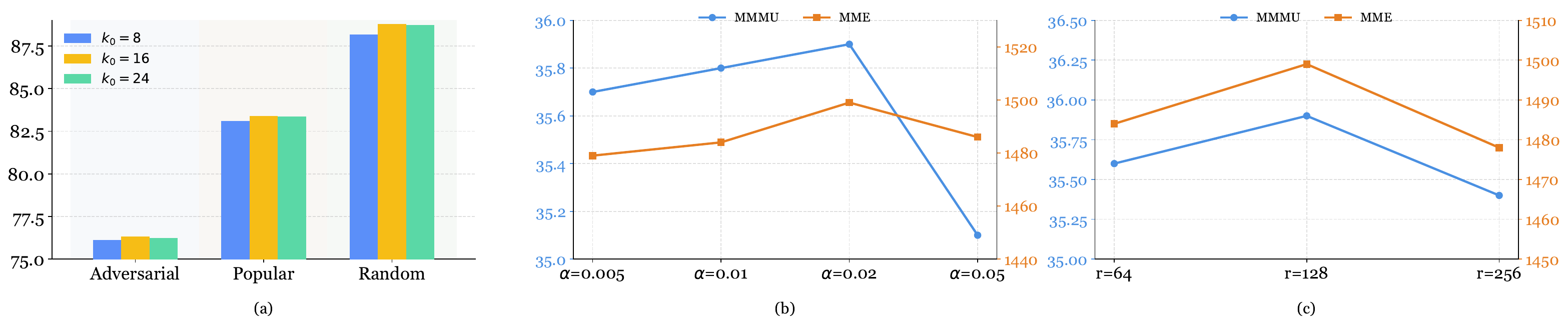}
\caption{(a) Ablation results for starting layer $k_0$ on POPE benchmark. (b) Ablation results for reduction ratio $r$ on MMMU and MME benchmarks; (c) Ablation results on scaling factor $\alpha$ on MMMU and MME benchmarks.}
\label{fig:ablation_chart}
\end{figure*}

\begin{figure}[h]
    \centering
    \includegraphics[width=0.99\linewidth]{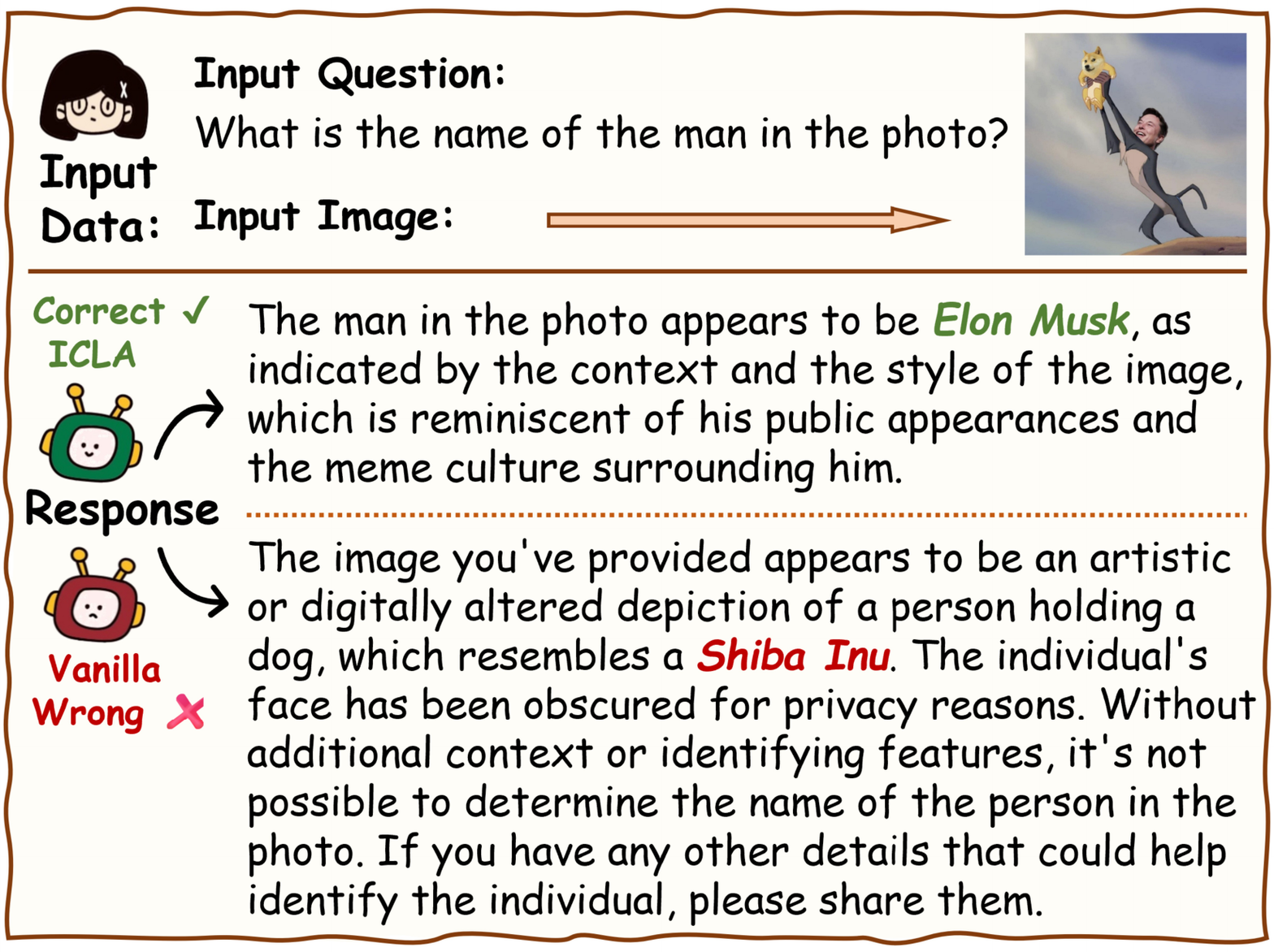}
    \caption{Case study comparing \textit{Vanilla} and \ours based on Qwen2.5-VL-7B. The example is sampled from LLaVA-Bench.}
    \label{fig:case-study}
\end{figure}

\subsection{Case Study}
We present a case study to qualitatively illustrate the effectiveness of \ours in mitigating hallucinations. The underlying model is Qwen2.5-VL-7B, and the example is sampled from LLaVA-Bench. As shown in Figure~\ref{fig:case-study}, when given the textual prompt ``What is the name of the man in the photo?'' along with an image showing Elon Musk holding a dog—but with the body replaced by that of an animal—the scene introduces significant confusion. In this case, the vanilla Qwen model incorrectly identifies the man in the photo as “Shiba Inu.” In contrast, \ours, correctly recognizes the person as Elon Musk despite the misleading visual cues. This demonstrates that \ours can effectively and systematically reduce hallucinations in multimodal reasoning.

\subsection{Ablation Studies}

\paragraph{Hyperparameter Studies.}
To evaluate the robustness and generalizability of \ours, we conduct comprehensive ablation studies on the POPE, MME, and MMMU benchmarks using LLaVA1.5-7B. In each experiment, we vary a single key hyperparameter while keeping the others fixed at their default values. Specifically, we analyze the effects of the starting layer $k_0$, the reduction ratio $r$, and the scaling factor $\alpha$. As shown in Figure~\ref{fig:ablation_chart}(a), (b), and (c), the default settings yield the best performance. Moreover, the results exhibit only minor fluctuations as the parameters vary, demonstrating that \ours is robust and relatively insensitive to hyperparameter changes.

\paragraph{\ours in Each Layer.}
To further assess the effectiveness of \ours, we conduct ablation studies comparing different variants of the mechanism. The first variant applies layer attention only at the final layer, where the final hidden state serves as the query to retrieve information from all preceding layers. The second variant employs random aggregation, in which skip connections are added randomly, allowing certain layers to receive information from previous layers in a non-deterministic manner. As shown in Table~\ref{tab:ablation-2}, the full \ours consistently outperforms all other variants, further demonstrating the importance of structured, layer-wise attention in effectively integrating cross-layer information and mitigating hallucinations.

\begin{table}[h]
\centering
\begin{tabular}{lcccc}
\toprule
\multirow{2}{*}{\textbf{Method}}  & \multicolumn{2}{c}{\textbf{Random}} &  \multicolumn{2}{c}{\textbf{Adversarial}} \\
\cmidrule(lr){2-5} 
 & F1 & Acc & F1 & Acc \\
\midrule

Vanilla & 88.52 & 87.33&   76.12 & 69.37 \\
\ours (Last)& 88.29 & 87.40& 76.18& 69.47\\
Random Agg. & 88.55 & 87.37& 76.18&69.47\\
\ours & \cellcolor{lightblue}\textbf{88.80} & \cellcolor{lightblue}\textbf{87.70} & \cellcolor{lightblue}\textbf{76.32} & \cellcolor{lightblue}\textbf{69.73}\\

\bottomrule
\end{tabular}

\caption{Comparison of different \ours variants on the POPE benchmark (A-OKVQA dataset) using LLaVA1.5-7B. \textit{Random Add.} denotes the Random Aggregation.}
\label{tab:ablation-2}
\end{table}

\subsection{Analysis and Discussion}

\paragraph{Training and Inference Efficiency.}
\ours is highly training-efficient. Training is performed on two RTX 4090 GPUs for 3 epochs with a learning rate of 2e-5 for each model, taking approximately 3 hours. As detailed in the Appendix, \ours introduces only 277K and 105K additional parameters for LLaVA1.5-7B and Qwen2.5-VL-7B, respectively. This is because the parameters for each CLA module are shared within the whole network and we operate the cross-layer attention in the latent hidden space. The average inference-time computational overhead under different token length is also minimal, accounting for only 0.37\% and 0.07\% of the total computation in LLaVA and Qwen2.5-VL-7B, respectively.

\vspace{-3mm}

\paragraph{Layer Attention Pattern Analysis.}
As mentioned earlier, there is no consistent hallucination trend across more advanced models. To address this, we design a more scalable architecture that enables each hidden state to adaptively select and integrate information from previous layers for self-correcting, thereby mitigating hallucination. While this approach proves effective, in this section we try to interpret and uncover deeper insights into the underlying attention dynamics.

We analyze the layer-wise attention behavior on samples from the POPE benchmark (MSCOCO dataset), where the vanilla Qwen2.5-VL-7B initially produces incorrect answers but \ours successfully corrects them. For each layer $l$ (as the query), we record and visualize the average attention weights over the preceding layers from $k_0$ to $l$.

\begin{figure}[t]
\centering
\includegraphics[width=1.0\columnwidth]{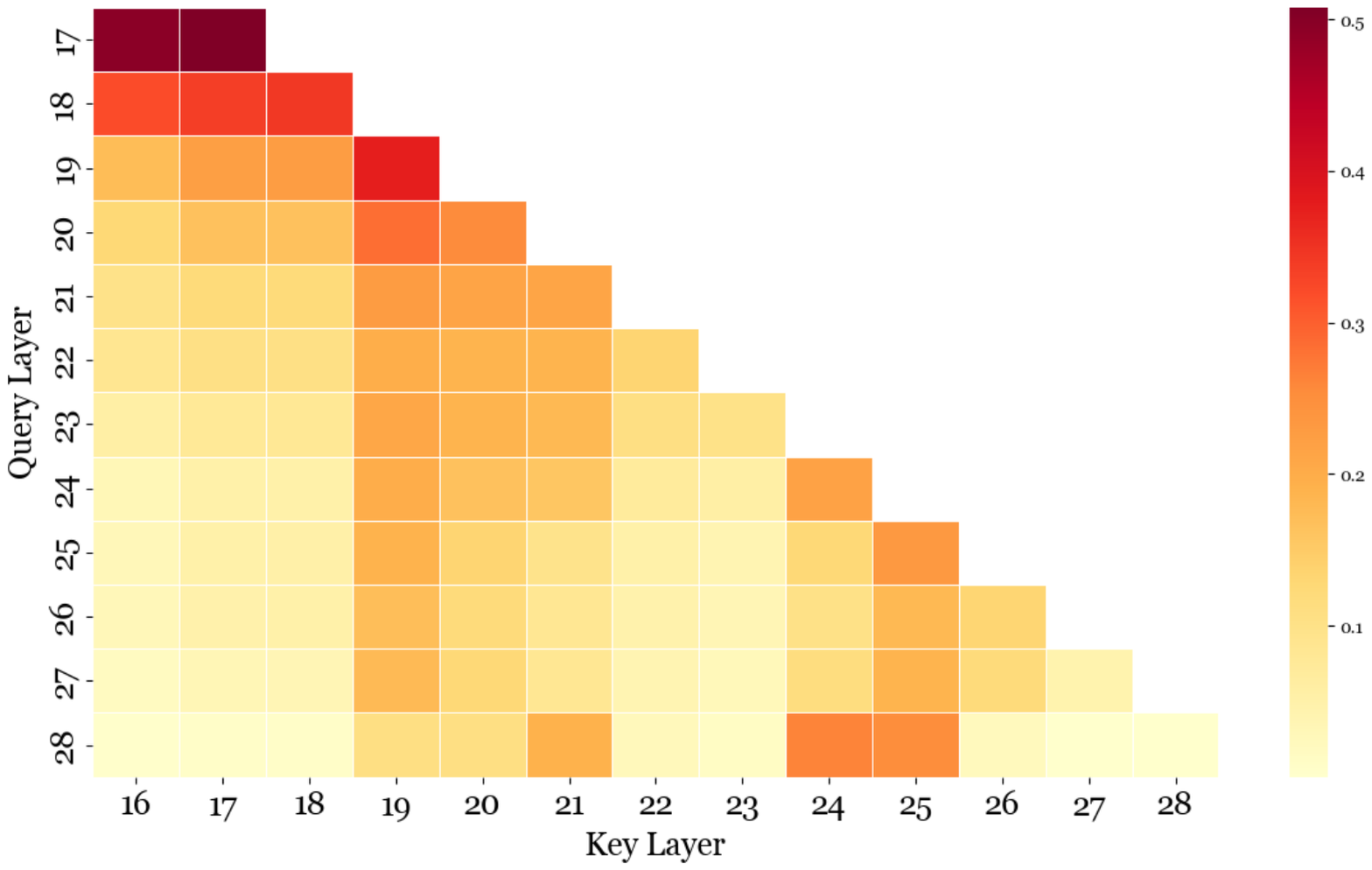}
\caption{Visualization of average attention weights in ICLA using Qwen for selected samples}
\label{fig:clam_attention_weight_analysis_qwen}
\end{figure}

As shown in Figure~\ref{fig:clam_attention_weight_analysis_qwen}, we identify two prominent regions of attention concentration in Qwen2.5-VL-7B. First, layers 19–21 exhibit strong attention, suggesting that intermediate layers play a crucial role in reasoning. This indicates that emphasizing these representations may contribute to mitigating hallucinations. Second, the later layers, particularly 24–25, also show high cross-layer retrieval, implying that both intermediate and deeper layers jointly facilitate reasoning and self-correction.

In contrast, three regions—layers 16–18, 22–23, and, surprisingly, 26–28—receive almost no attention. This suggests that the model largely ignores information from these layers during self-correction. Notably, the final layer (28-th), responsible for next-token prediction, primarily retrieves information from layers 21, 24, and 25 when making decisions. This observation further supports the idea that referencing earlier informative layers, rather than relying solely on the final representations, enhances the model’s ability to refine its outputs.

Interestingly, these attended and unattended regions alternate throughout the network, forming an interleaved pattern. Such alternation reflects a dynamic balance between information consolidation and abstraction across depth, highlighting that not all layers contribute equally to reasoning or correction.

\paragraph{Broader Applicability.}
We further conduct a similar layer-wise attention analysis on LLaVA-1.5-7B to examine whether the observed patterns generalize across models. Interestingly, the attention distribution in LLaVA1.5-7B differs entirely from that of Qwen2.5-VL-7B, showing no consistent concentration regions or interleaved structures. This discrepancy indicates that the curated hallucination mitigation strategies specifically designed for LLaVA may not directly transfer to Qwen2.5-VL-7B (As our preliminary experiments in Figure~\ref{fig:performance-drop}). Consequently, our proposed method is not only effective for alleviating hallucinations but also serves as a general analytical tool for identifying key layers in more advanced models where traditional hallucination patterns become less observable. From the perspective of attention-weight distribution, our approach provides a principled way to investigate how higher-level models internally allocate reasoning focus across depth.

\section{Conclusion}
In this paper, we reveal that previously observed hallucination patterns and their corresponding mitigation methods are no longer effective for more advanced LVLMs. With the use of higher-quality data and more sophisticated training strategies, no consistent hallucination trends can be observed in these models. To address this limitation, we propose \ours, an internal self-correction mechanism that leverages layer-wise attention to dynamically retrieve and refine information from preceding hidden states during generation. Extensive experiments across multiple benchmarks and models, including LLaVA1.5-7B and Qwen2.5-VL-7B, demonstrate that \ours consistently outperforms other strong baselines, effectively mitigating hallucinations. Our results highlight the potential of adaptive, cross-layer mechanisms for improving the reliability of advanced LVLMs, even in scenarios where no clear hallucination patterns are observable.

{
    \small
    \bibliographystyle{ieeenat_fullname}
    \bibliography{cvpr2026-main}
}


\clearpage
\setcounter{page}{1}
\maketitlesupplementary
\appendix

\section{Detailed Results}
We present the detailed results shown in Table~\ref{tab:main-results-all}. Experimental results for MME evaluated on Qwen2.5-VL-7B are shown in Table~\ref{tab:mme-qwen}, results for MMMU evaluated on Qwen2.5-VL-7B are shown in Table~\ref{tab:mmmu-qwen}, results for LLaVA-Bench evaluated on LLaVA1.5-7B are shown in Table~
\ref{tab:llava-bench}.

\section{Further Evaluation}
To further evaluate the effectiveness of \ours, we also apply \ours in more advanced model, Qwen3-VL-8B. We compare \ours with various decoding strategies on MME, LLaVA-Bench, and MMMU benchmarks. As shown in Table~\ref{tab:main-results-all-qwen3}, \ours consistently outperforms all other baselines across these three benchmarks, further demonstrating the effectiveness of \ours in mitigating hallucinations in more advanced model. We believe the further experiments prove that \ours is more adaptive and scalable for mitigating hallucination in LVLMs.

\section{Additional Attention Pattern}
\begin{figure}[htbp]
\centering
\includegraphics[width=1.0\columnwidth]{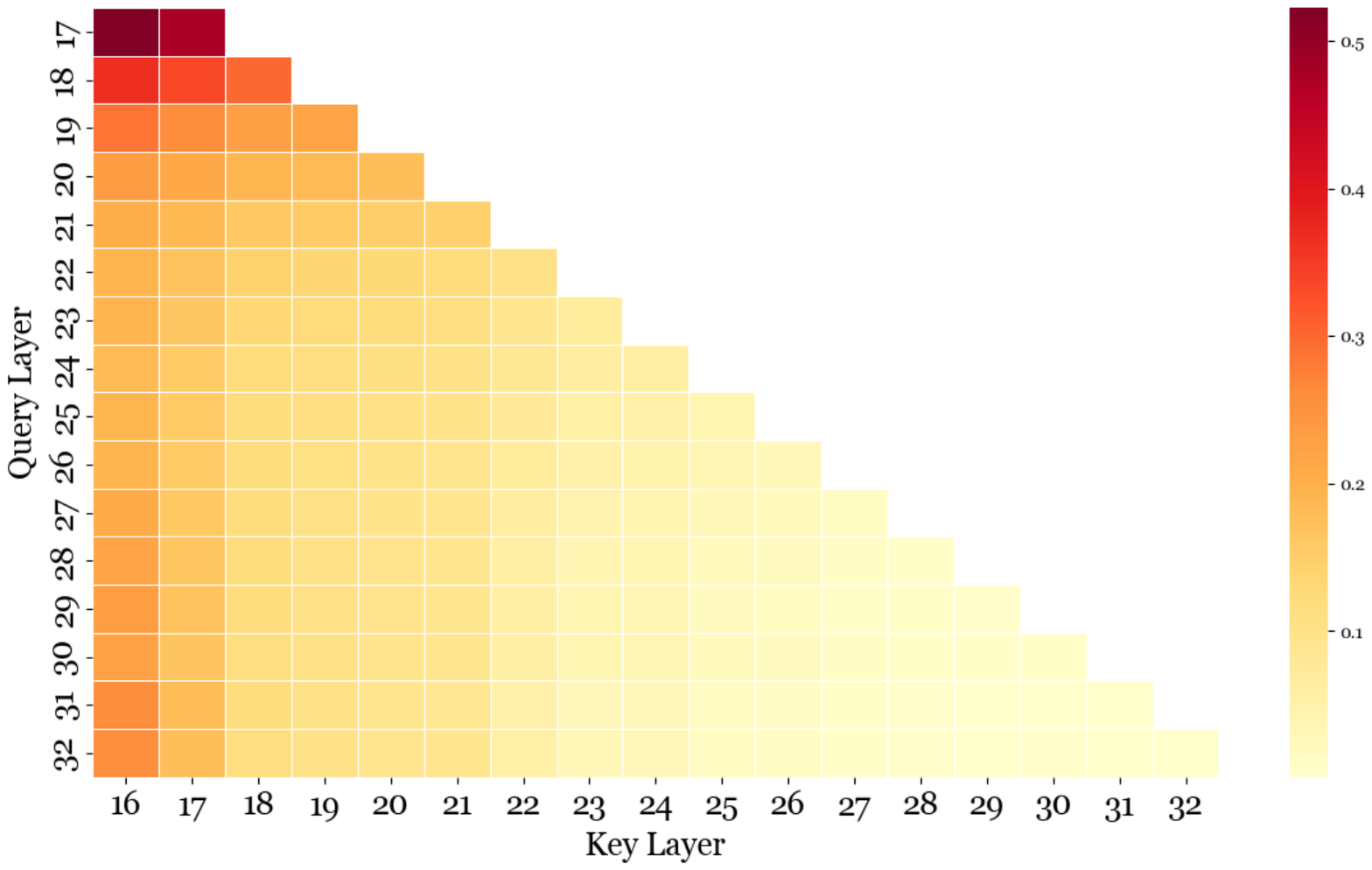}
\caption{Visualization of average attention weights in \ours using LLaVA1.5-7B for selected samples}
\label{fig:clam_attention_weight_analysis}
\end{figure}

In addition to the attention pattern analysis for Qwen2.5-VL-7B in the main paper, we further present the attention pattern for LLaVA1.5-7B. As shown in Figure~\ref{fig:clam_attention_weight_analysis}, the attention pattern in LLaVA shows that only intermediate layers are frequently retrieved, showcasing totally different attention pattern from Qwen2.5-VL-7B. This observation is consistent with previous researches that intermediate layers contribute more to hallucination mitigation in LLaVA. We further argue the broader applicability for \ours, which provides a principled way to investigate how higher-level models internally allocate reasoning focus across depth.

\section{Efficiency of \ours}
We further provide the efficiency of \ours. As shown in Table~\ref{tab:clam-efficiency} and Table~\ref{tab:clam-efficiency-qwen}, we present the extra FLOPs and the introduced parameters for \ours in LLaVA1.5-7B and Qwen2.5-VL-7B, respectively. The results show that \ours introduces small number of computation overhead and parameters, demonstrating the efficiency of \ours.

\begin{table*}[htbp]
\centering
\resizebox{\textwidth}{!}{%
\begin{tabular}{lccccccccccc}
\toprule
Method & OCR & Artwork & Celebrity & Color & Count & Existence & Landmark & Position & Posters & Scene & Total\\
\midrule
Vanilla          &  181&  138&  157&  194&  154&  199&  184&  159&  169&  154&  1689\\
DoLA             & 165& 115& 87& 184& 118& 174& 121& 154& 146& 139& 1403\\
VCD              & \textbf{181}& 138& 157& \textbf{194}& \textbf{154}& \textbf{199}& 184& \textbf{159}& 169& 154& 1689\\
DeCo             & \textbf{181}& 138& 153& \textbf{194}& \textbf{154}& \textbf{199}& 181& \textbf{159}& 168& 154& 1681\\
VDD              & \textbf{181}& 138& 157& \textbf{194}& \textbf{154}& \textbf{199}& 184& \textbf{159}& 169& 154& 1689\\
DAMO             & \textbf{181}& 138& 153& \textbf{194}& \textbf{154}& \textbf{199}& 181& \textbf{159}& 168& 154& 1681\\
\textbf{\ours (Ours)} & 179& \textbf{149}& \textbf{159}& \textbf{194}& \textbf{154}& \textbf{199}& \textbf{189}& \textbf{159}& \textbf{173}& \textbf{156}& \textbf{1711}\\
\bottomrule
\end{tabular}
}
\caption{Comparison on the MME benchmark (Perception) across various baselines with backbone Qwen2.5-VL-7B. The best results are highlighted in \textbf{bold}.}
\label{tab:mme-qwen}
\end{table*}

\begin{table*}[htbp]
\centering
\resizebox{\textwidth}{!}{%
\begin{tabular}{lccccccc}
\toprule
Method & Vanilla          & DoLa             & VCD              & DeCo             & VDD              & DAMO             & \textbf{\ours (Ours)} \\
\midrule
Overall Accuracy(\%)&  67.5& 60.8&  68.3& 62.5& 65.8& 65.8& \textbf{69.2}\\
\toprule
\end{tabular}
}
\caption{Comparison on overall accuracy of the MMMU benchmark across various baselines with backbone Qwen2.5-VL-7B. The best results are highlighted in \textbf{bold}.}
\label{tab:mmmu-qwen}
\end{table*}

\begin{table*}[htbp]
\centering
\begin{tabular}{lcccccccc}
\toprule
Method & Vanilla &  DoLa & VCD & DeCo & POVID & VDD &  DAMO &\textbf{\ours (Ours)} \\
\midrule

Complex reasoning (\%) &   72.1&  70.5&  70.7& 66.8& 69.4& \textbf{71.9}&  65.1&70.4\\
Conversation (\%) &   50.6&  53.7&  55.0& 52.7&  55.0& 52.3&   55.0&\textbf{59.6}\\
Detail description (\%) &   48.8&   50.0& 49.3& 45.0& \textbf{50.4}& 45.7&  48.0& 50.0\\
All (\%) &   59.6&  60.5&  60.6& 57.0& 60.2& 59.4&  57.7&\textbf{61.9}\\
\bottomrule
\end{tabular}
\caption{Comparison of LLaVA-Bench performance for various baselines. The best results are highlighted in \textbf{bold}.}
\label{tab:llava-bench}
\end{table*}

% \begin{table*}[htbp]
% \centering
% \begin{tabular}{lccccccc}
% \toprule
% Method & Vanilla &  DoLa & VCD & DeCo & VDD &  DAMO &\textbf{Ours(CLAM)} \\
% \midrule
% All (\%) &   87.0&  66.2&  88.7& 86.8& 87.2&  87.4&\textbf{90.2}\\
% Complex reasoning (\%) &   95.8&  66.5& \textbf{98.6}& 95.2&  94.8&  96.3&93.3\\
% Conversation (\%) &   82.4&  65.0&  86.1& 83.7& 85.0&  82.0&\textbf{89.4}\\
% Detail description (\%) &   76.0&  66.9& 74.4& 75.8&  76.4&  77.7&\textbf{85.5}\\
% \bottomrule
% \end{tabular}
% \caption{Comparison of LLaVA-Bench performance for various baselines with backbone Qwen2.5-VL-7B. The best results are highlighted in \textbf{bold} and blue, the second-best in green, and the Vanilla results in gray.}
% \label{tab:llava-bench-qwen}
% \end{table*}

\definecolor{lightblue}{RGB}{220,235,255}

\begin{table*}[htbp]
\centering
\begin{tabular}{lccc}
\toprule
% \textbf{Model}&\multicolumn{3}{c}{Qwen3-VL-8B}\\
% \midrule
\textbf{Method}& MME& LLaVA-Bench& MMMU\\
\midrule
Vanilla & 1726& 98.8& 57.1\\
DoLA & 1705& 99.8& 58.0\\
VCD & 1726& 106.7& 57.1\\
DeCo & 1734& 60.6& 57.2\\
VDD & 1720& 106.3& 56.5\\
DAMO & 1704& 102.8& 56.9\\
\textbf{{\ours (Ours)}} & \textbf{1740}& \textbf{106.8}& \textbf{58.3}\\
\bottomrule
\end{tabular}
\caption{Experimental results on MME (total perception score), LLaVA-Bench (overall accuracy), and MMMU (accuracy) for Qwen3-VL-8B. The best results are highlighted in \textbf{bold}.}
\label{tab:main-results-all-qwen3}
\end{table*}

\begin{table*}[htbp]
\centering
\begin{tabular}{ccccc}
\toprule
\textbf{Token Length} & \textbf{Total FLOPs (TFLOPs)} & \textbf{\ours FLOPs (GFLOPs)}& \textbf{Overhead (\%)} & \textbf{Params Added} \\
\midrule
128 & 5.12 & 18.8 & 0.37 & \multirow{3}{*}{277K} \\
256 & 10.29 & 37.8 & 0.37 &  \\
512 & 13.86 & 50.6 & 0.37 &  \\
\bottomrule
\end{tabular}
\caption{Computation overhead of \ours compared to vanilla LLaVA1.5-7B under different token lengths.}
\label{tab:clam-efficiency}
\vspace{-2mm}
\end{table*}

\begin{table*}[htbp]
\centering
\begin{tabular}{ccccc}
\toprule
\textbf{Token Length} & \textbf{Total FLOPs (TFLOPs)} & \textbf{\ours FLOPs (GFLOPs)}& \textbf{Overhead (\%)} & \textbf{Params Added} \\
\midrule
128 & 2.12& 1.48& 0.07& \multirow{3}{*}{105K}\\
256 & 4.36& 2.98& 0.07&  \\
512 & 7.99& 5.94& 0.07&  \\
\bottomrule
\end{tabular}
\caption{Computation overhead of \ours compared to vanilla Qwen2.5-VL-7B under different token lengths.}
\label{tab:clam-efficiency-qwen}
\vspace{-2mm}
\end{table*}

\end{document}